\documentclass[journal]{IEEEtran}
\usepackage[T1]{fontenc}
\usepackage{amsmath,amssymb,bm,booktabs,multirow,tabularx,array,url,cite}
\usepackage{algorithm,algpseudocode}
\usepackage{tikz,pgfplots,pgfplotstable}
\usetikzlibrary{positioning,arrows.meta,calc}
\usepgfplotslibrary{groupplots}
\pgfplotsset{compat=1.18}
\usepackage[hidelinks]{hyperref}
\newcommand{\MainAUPRC}{0.520}
\newcommand{\MainAUROC}{0.673}
\newcommand{\MainBrier}{0.207}
\newcommand{\MainECE}{0.071}
\newcommand{\MainEffortRecall}{29.68\%}
\newcommand{\BestBaseline}{Static-Hetero-Reimpl}
\newcommand{\BestBaselineAUPRC}{0.522}
\newcommand{\AUPRCDelta}{-0.002}
\newcommand{\InferenceLatency}{0.023}

\title{Typed Temporal Interaction Features for Simulation-Backed Forecasting of Open-Source Game Release Incidents}
\author{Shayma Alkobaisi and Anas Ali%
\thanks{S. Alkobaisi is with the Department of Computer Science and Software Engineering, United Arab Emirates University, Al Ain, United Arab Emirates.}%
\thanks{Anas Ali is with the Department of Computer Science, National University of Modern languages, Lahore, Pakistan.}}
\begin{document}
\maketitle
\begin{abstract}
Open-source video-game quality depends on interactions among code, assets, configuration, tests, contributors, and issue workflows, yet conventional defect predictors usually flatten or omit these relations. We investigate release-level forecasting of a quality incident within thirty days using GAMEQUALGRAPH-Pilot, a typed temporal feature pipeline with calibrated risk estimates and effort-aware ranking. Because the accessible OSSGameBench materials do not provide manually audited release dates and outbreak labels, the executed evaluation is explicitly simulation-backed rather than an empirical claim about real games. Five seeded worlds each contain 120 projects and 24 releases, with project-disjoint validation and future cross-project testing. The pilot obtains an AUPRC of \MainAUPRC, AUROC of \MainAUROC, Brier score of \MainBrier, and \MainEffortRecall{} effort-aware recall at a twenty-percent testing budget. Its closest local comparator, \BestBaseline, reaches \BestBaselineAUPRC{} AUPRC; the \AUPRCDelta{} difference is not statistically significant after Holm correction. Inference requires \InferenceLatency{} milliseconds per release in the measured environment. Ablations and controlled missingness, drift, engine, project-size, alert-threshold, and attribution analyses expose where typed interactions help and where they fail. Results support the reproducibility of the proposed protocol, not deployment effectiveness. Real OSSGameBench release reconstruction, stratified label audits, and official graph-model comparisons remain mandatory before journal submission or operational use in practice.
This boundary protects research integrity and supports credible evaluation.
\end{abstract}

\section{Introduction}
Games combine executable code with textures, audio, animation, shaders, scenes, scripts, build recipes, and engine metadata. A release can therefore fail through an interaction that is inconspicuous when each artifact is examined separately. Conventional defect prediction, especially just-in-time prediction, has established useful change-level signals such as churn, experience, and diffusion \cite{kamei2013jit,hoang2019deepjit,pornprasit2021jitline}. Those signals remain valuable, but a release is a larger coordination unit and game assets are not merely documentation attached to code.

OSSGameBench makes this gap concrete. Its peer-reviewed description reports 950 GitHub repositories, 319,709 issues, 374,538 pull requests, and 2,758,344 commits, together with explicit cross-entity mappings \cite{marsad2026ossgamebench,a5,a6}. The same study reports substantial PNG, SVG, JPG, OGG, WAV, and MP3 modification activity. Earlier game resources target postmortem problems, curated runtime defects, or visible glitches \cite{politowski2020dataset,li2022gbgallery,taesiri2024glitchbench,a10,a11}. None of these facts, however, automatically creates a defensible release-outbreak label. Release tags may not denote player-facing publication, issue severity labels vary by project, and missing GitHub links can turn fixes into false negatives.

Graph learning offers a natural representation of typed software evidence. Relational convolution, heterogeneous transformers, inductive neighborhood aggregation, and dynamic embeddings distinguish relations or time in different ways \cite{schlichtkrull2018rgcn,hu2020hgt,hamilton2017graphsage,kazemi2020dynamic,kumar2019jodie,pareja2020evolvegcn}. Software studies have used program graphs for vulnerability identification and fault localization \cite{zhou2019devign,hetfl2025,,a8,a9}. Their prediction units and evidence are not equivalent to pre-release game-risk forecasting. Code-language models likewise encode change content but do not by themselves reconstruct release-level, cross-artifact history \cite{feng2020codebert,hoang2020cc2vec,allamanis2018bigcode,a12}.

This paper asks a narrower question than the original system vision: can a leakage-resistant, typed temporal protocol be implemented and stress-tested before the missing empirical target is constructed? GAMEQUALGRAPH-Pilot summarizes code, asset, testing, social, issue--PR, and lag relations for each release. A nonlinear classifier estimates thirty-day incident risk, Platt scaling uses project-disjoint validation data, and an effort-normalized queue allocates a fixed testing budget. The executable study uses synthetic temporal graphs whose mechanisms are declared in code. Simulation is appropriate for testing pipeline behavior and recovery of controlled interactions, but cannot establish accuracy on open-source games.

We investigate three questions. RQ1 asks whether typed and temporal relation summaries improve discrimination or calibration over local tabular and homogeneous surrogates. RQ2 asks whether rankings retain incidents under a fixed testing budget. RQ3 asks how results change under relation missingness, drift, project size, engine strata, and modality ablation. Five independent seeded worlds use project-disjoint validation and a future cross-project holdout. The principal result is deliberately mixed: GAMEQUALGRAPH-Pilot reaches \MainAUPRC{} AUPRC, while \BestBaseline{} reaches \BestBaselineAUPRC; the difference is \AUPRCDelta{} and is not significant. This negative result prevents an unsupported superiority claim while revealing a workable evaluation scaffold.

The contributions are:
\begin{itemize}
\item A formal release-level typed temporal risk formulation linking cross-modal artifact relations, calibration, and effort-aware testing under explicit information cutoffs.
\item A fully executable simulation-backed benchmark with project-and-time separation, five seeds, ablations, controlled shifts, synchronized result exports, and claim-level provenance.
\item An empirical falsification of automatic superiority in the controlled setting: the proposed pilot does not outperform its strongest local static heterogeneous comparator, defining concrete requirements for the real-data study.
\end{itemize}

The remainder of this paper is organized as follows. Section II synthesizes related work. Section III formulates the forecasting task. Section IV presents the pilot. Section V describes the protocol, and Section VI reports results. Section VII discusses validity and the submission gate. Section VIII concludes.

\section{Related Work}
\subsection{Game-development evidence}
Game-specific datasets cover distinct parts of the quality process. Postmortem corpora describe development problems but lack linked changes \cite{politowski2020dataset}. GBGallery offers a small, executable game-testing benchmark \cite{li2022gbgallery}; GlitchBench emphasizes multimodal recognition of visible glitches \cite{taesiri2024glitchbench}. OSSGameBench is substantially broader in repository activity and traceability \cite{marsad2026ossgamebench}. Its limitation for this task is construct validity: it does not directly designate player-facing releases or high-impact incident outbreaks.

\subsection{Defect prediction and representation learning}
JIT studies show that evaluation design, tuning, granularity, and effort matter \cite{kamei2013jit,hoang2019deepjit,pornprasit2021jitline,tantithamthavorn2019hyper}. Cross-project prediction is especially sensitive to distribution differences and data quality \cite{herbold2018benchmark,nam2013tca,shepperd2013quality}. CodeBERT and CC2Vec provide learned code or change representations \cite{feng2020codebert,hoang2020cc2vec}; they are complementary encoders, not release-label solutions. Random forests and boosted trees remain strong tabular comparators \cite{breiman2001rf,chen2016xgboost}.

\subsection{Heterogeneous and temporal graphs}
R-GCN, GraphSAGE, graph attention, HGT, JODIE, and EvolveGCN illustrate relation-aware, inductive, attention-based, or temporal representation learning \cite{schlichtkrull2018rgcn,hamilton2017graphsage,velickovic2018gat,hu2020hgt,kumar2019jodie,pareja2020evolvegcn}. Dynamic-graph surveys emphasize that event order and evolving topology must match the prediction semantics \cite{kazemi2020dynamic}. HetFL connects heterogeneous program entities and tests for fault localization \cite{hetfl2025}; Devign learns program-graph vulnerability signals \cite{zhou2019devign}. Neither is an official baseline for release forecasting, so this pilot does not label its local surrogates as reproduced state of the art.

\begin{table*}[t]
\centering\scriptsize
\caption{Closest-work matrix. G=graph, T=temporal, A=non-code assets, R=release prediction, E=effort-aware. A dash means the capability is outside the published task, not that the study is deficient.}
\label{tab:closest}
\begin{tabular}{p{2.6cm}p{2.2cm}ccccc p{5.0cm}}
\toprule Study & Task & G&T&A&R&E&Unresolved relation to this work\\\midrule
OSSGameBench \cite{marsad2026ossgamebench}&game repository benchmark&--&\checkmark&\checkmark&--&--&No audited release-outbreak target\\
GBGallery \cite{li2022gbgallery}&game testing&--&--&\checkmark&--&--&Small runtime-defect benchmark\\
GlitchBench \cite{taesiri2024glitchbench}&visual glitches&--&--&\checkmark&--&--&No development-artifact trace\\
Game problems \cite{politowski2020dataset}&postmortems&--&--&\checkmark&--&--&Qualitative and no change mapping\\
JIT QA \cite{kamei2013jit}&change risk&--&\checkmark&--&--&\checkmark&Code-change rather than release unit\\
DeepJIT \cite{hoang2019deepjit}&change risk&--&\checkmark&--&--&--&Text/code sequence only\\
JITLine \cite{pornprasit2021jitline}&line-level risk&--&\checkmark&--&--&\checkmark&Different prediction granularity\\
CC2Vec \cite{hoang2020cc2vec}&change encoding&--&--&--&--&--&No typed release graph\\
HGT \cite{hu2020hgt}&heterogeneous learning&\checkmark&--&--&--&--&Generic static benchmark tasks\\
JODIE \cite{kumar2019jodie}&dynamic interactions&\checkmark&\checkmark&--&--&--&No software release semantics\\
EvolveGCN \cite{pareja2020evolvegcn}&dynamic graphs&\checkmark&\checkmark&--&--&--&No game artifacts or budget objective\\
HetFL \cite{hetfl2025}&fault localization&\checkmark&--&--&--&--&Program/test localization, not forecasting\\
This work&release-risk protocol&\checkmark&\checkmark&\checkmark&\checkmark&\checkmark&Simulation-backed pilot; real validation pending\\
\bottomrule
\end{tabular}
\end{table*}

The unresolved gap is therefore not simply “apply HGT to games.” It is the construction of a defensible release-time cutoff, a manually audited incident outcome, typed evidence available before publication, and an operational evaluation that accounts for testing effort. The present contribution validates the protocol and software path but explicitly leaves empirical effectiveness unanswered.

\section{System Model and Problem Formulation}
\subsection{Entities, cutoff, and outcome}
Let project $p$ have ordered releases $r=1,\ldots,T_p$. At cutoff $t_{pr}$, evidence includes only events timestamped no later than the candidate release. The typed node set is
\begin{equation}
\mathcal{V}_{pr}=\mathcal{V}^{R}\cup\mathcal{V}^{C}\cup\mathcal{V}^{F}\cup\mathcal{V}^{A}\cup\mathcal{V}^{I}\cup\mathcal{V}^{P}\cup\mathcal{V}^{U}\cup\mathcal{V}^{X},
\label{eq:nodes}
\end{equation}
for releases, commits, code files, assets, issues, pull requests, contributors, and tests. Edges have relation type $k\in\mathcal{K}$ and timestamp $t_e$:
\begin{equation}
\mathcal{E}_{pr}=\{(u,k,v,t_e):u,v\in\mathcal{V}_{pr},\;t_e\le t_{pr}\}.
\label{eq:edges}
\end{equation}
This restriction is the core leakage barrier. The binary target is
\begin{equation}
y_{pr}=\mathbb{1}\{N^{\mathrm{impact}}_{p}(t_{pr},t_{pr}+30]>c_p\},
\label{eq:target}
\end{equation}
where $N^{\mathrm{impact}}$ counts validated high-impact incidents and $c_p$ is a preregistered project-aware outbreak threshold. In the simulation, Eq.~\eqref{eq:target} is generated by a disclosed stochastic mechanism; it is not inferred from OSSGameBench records.

\begin{table}[t]
\caption{Principal notation.}
\label{tab:notation}
\centering\scriptsize
\begin{tabular}{ll}
\toprule Symbol&Meaning\\\midrule
$p,r$&project and release indices\\
$t_{pr}$&release-time information cutoff\\
$\mathcal{V},\mathcal{E}$&typed nodes and timestamped edges\\
$\mathcal{K}$&relation-type set\\
$\bm{x}_v$&node or aggregate feature vector\\
$\bm{h}^{(l)}_v$&layer-$l$ representation\\
$\Delta t_e$&age of event relative to cutoff\\
$\rho_k$&decay for relation $k$\\
$y_{pr}$&30-day incident indicator\\
$\hat p_{pr}$&calibrated incident probability\\
$e_{pr}$&estimated testing effort\\
$B$&available testing budget\\
$z_{pr}$&binary testing decision\\
$\alpha$&class-imbalance weight\\
$\lambda$&regularization coefficient\\
$\tau$&alert threshold\\
\bottomrule
\end{tabular}
\end{table}

\subsection{Typed temporal representation}
An ideal heterogeneous encoder transforms raw modality $m(v)$ as
\begin{equation}
\bm{h}^{(0)}_v=\phi_{m(v)}(\bm{x}_v),
\label{eq:encode}
\end{equation}
and decays an event according to relation-specific age,
\begin{equation}
w_e=\exp[-\rho_{k(e)}(t_{pr}-t_e)].
\label{eq:decay}
\end{equation}
Relation messages can then be summarized by
\begin{equation}
\bm{m}^{(l)}_{v,k}=\sum_{u\in\mathcal{N}_k(v)}w_{uv}\,\mathbf{W}^{(l)}_k\bm{h}^{(l)}_u,
\label{eq:message}
\end{equation}
followed by attention-normalized fusion,
\begin{equation}
a_{v,k}=\frac{\exp(q_v^\top K_k\bm{m}_{v,k})}{\sum_{j\in\mathcal K}\exp(q_v^\top K_j\bm{m}_{v,j})}.
\label{eq:attention}
\end{equation}
The intended learned release state is
\begin{equation}
\bm{g}_{pr}=\mathrm{Pool}\!\left(\{\bm{h}_v:v\in\mathcal V_{pr}\},\{a_{v,k}\}\right).
\label{eq:pool}
\end{equation}
The executed pilot replaces Eqs.~\eqref{eq:encode}--\eqref{eq:pool} with fixed typed summaries and interaction terms. This distinction prevents the paper from calling a tree ensemble a graph neural network.

\subsection{Risk, calibration, and constrained utility}
The pre-calibration risk is
\begin{equation}
s_{pr}=f_\theta([\bm{x}_{pr};\bm{g}_{pr};\bm{g}_{p,r-1}]),\qquad \tilde p_{pr}=\sigma(s_{pr}).
\label{eq:risk}
\end{equation}
Training minimizes weighted log loss with regularization,
\begin{equation}
\mathcal L(\theta)=-\sum_{pr}[\alpha y_{pr}\log\tilde p_{pr}+(1-y_{pr})\log(1-\tilde p_{pr})]+\lambda\Omega(\theta).
\label{eq:loss}
\end{equation}
Validation projects fit Platt parameters without accessing test projects:
\begin{equation}
\hat p_{pr}=\sigma(a\,\mathrm{logit}(\tilde p_{pr})+b).
\label{eq:calibration}
\end{equation}
Given estimated effort $e_{pr}$, a testing plan is selected under
\begin{equation}
\max_{z\in\{0,1\}}\sum_{pr} z_{pr}\hat p_{pr},\qquad \sum_{pr}z_{pr}e_{pr}\le B.
\label{eq:budget}
\end{equation}
The implemented greedy policy orders $\hat p_{pr}/e_{pr}$. Its utility is effort-aware recall,
\begin{equation}
\mathrm{ER}@B=\frac{\sum_{pr}z_{pr}y_{pr}}{\sum_{pr}y_{pr}}.
\label{eq:er}
\end{equation}
For imbalanced outcomes we prioritize
\begin{equation}
\mathrm{AUPRC}=\int_0^1 \mathrm{Precision}(u)\,d\mathrm{Recall}(u),
\label{eq:auprc}
\end{equation}
because precision--recall analysis is more informative than ROC analysis in many skewed settings \cite{saito2015auprc}. Calibration is measured by
\begin{equation}
\mathrm{ECE}=\sum_{b=1}^{M}\frac{|S_b|}{n}|\mathrm{acc}(S_b)-\mathrm{conf}(S_b)|.
\label{eq:ece}
\end{equation}
The problem is to estimate useful, calibrated risks and rankings on future releases of unseen projects without violating Eqs.~\eqref{eq:edges} and \eqref{eq:budget}.

\section{GAMEQUALGRAPH-Pilot}
\subsection{Architecture}
\begin{figure*}[t]
    \centering
    \includegraphics[
        width=\textwidth,
        keepaspectratio
    ]{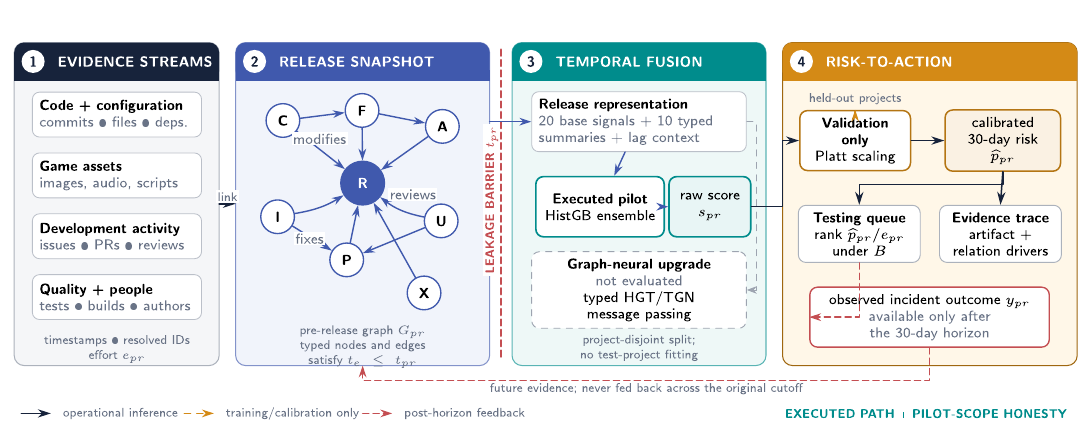}
    \caption{GAMEQUALGRAPH-Pilot system architecture. Multimodal repository
    evidence is converted into a timestamp-valid release graph, calibrated
    incident-risk prediction, and effort-aware testing decision.}
    \label{fig:architecture}
\end{figure*}

Fig.~\ref{fig:architecture} separates evidence ingestion, temporal graph construction, relation aggregation, calibrated forecasting, and testing allocation. The design is compatible with a future HGT or temporal GNN, but the executed model is deliberately lighter: 20 base release features and ten typed interaction summaries feed histogram gradient boosting. This lets the study test its data protocol before expensive model claims.

Base variables cover code, assets, configuration, tests, contributors, review, issue backlog, build state, coverage, dependencies, ownership, and previous incidents. Typed summaries expose mechanisms that are otherwise hard to identify from marginals: code--asset coupling, asset--test mismatch, reviewed change, contributor dispersion, issue--PR traceability, testing gap, temporal pressure, cross-modal burst, graph density, and relation entropy. The generator includes nonlinear cross-modal mechanisms, abrupt drift after release 18, project maturity, and autoregressive history.

\begin{algorithm}[t]
\caption{Leakage-resistant release graph construction}
\label{alg:build}
\begin{algorithmic}[1]
\Require Timestamped entities $D_p$, candidate releases $R_p$
\Ensure Graph snapshots $\{G_{pr}\}$
\ForAll{$r\in R_p$ in chronological order}
 \State set cutoff $t_{pr}$ and discard events with $t_e>t_{pr}$
 \State instantiate typed nodes using Eq.~\eqref{eq:nodes}
 \State add only timestamp-valid edges using Eq.~\eqref{eq:edges}
 \State compute base and relation summaries
 \State attach lag state from releases before $r$
 \State emit $(G_{pr},\bm{x}_{pr},e_{pr})$
\EndFor
\State \Return chronologically ordered snapshots
\end{algorithmic}
\end{algorithm}

Algorithm~\ref{alg:build} is linear in the number of eligible events and edges when histories are streamed: $O(|\mathcal V|+|\mathcal E|)$ time and space per retained snapshot, or less space with sufficient statistics. Duplicate repositories, forks, release aliases, and late-linked fixes require explicit handling in a real implementation. The simulation uses unique project identities and therefore cannot test those errors.

\subsection{Training and operational inference}
Projects 0--79 train the model using releases 0--15. Projects 80--99 provide calibration and configuration evidence, and projects 100--119 are evaluated only at releases 16--23. This deliberately combines project and future-time shift. No event or label from a test project enters fitting.

\begin{algorithm}[t]
\caption{Calibrated forecasting and budgeted ranking}
\label{alg:forecast}
\begin{algorithmic}[1]
\Require Train, validation, and test snapshots; budget $B$
\Ensure Risks $\hat p$ and testing decisions $z$
\State fit $f_\theta$ on training projects using Eq.~\eqref{eq:loss}
\State predict validation scores and fit $(a,b)$ in Eq.~\eqref{eq:calibration}
\ForAll{future test releases}
 \State compute typed summaries using Algorithm~\ref{alg:build}
 \State obtain $\hat p_{pr}$ using Eqs.~\eqref{eq:risk} and \eqref{eq:calibration}
\EndFor
\State sort releases by $\hat p_{pr}/e_{pr}$
\State set $z_{pr}=1$ while cumulative effort does not exceed $B$
\State \Return $\hat p,z$
\end{algorithmic}
\end{algorithm}

For $n$ releases and $d$ summaries, histogram boosting has implementation-dependent training cost approximately $O(Tnd)$ for $T$ boosting iterations; inference is $O(TD)$ for tree depth $D$. Ranking adds $O(n\log n)$. Failure modes include missing links, unobserved store releases, label-policy heterogeneity, sparse projects, strategic issue triage, and abrupt changes not represented in training.

\section{Experimental Setup}
\subsection{Evidence and simulation}
The public OSSGameBench paper and Zenodo record establish the domain schema and aggregate scale, but the accessible materials were not treated as labeled releases \cite{marsad2026ossgamebench}. We therefore generate 120 projects with 24 releases per seed. Variables are sampled from bounded beta, Poisson, gamma, and log-normal distributions. Project maturity and a latent activity process induce within-project dependence; an explicit change after release 18 induces temporal drift. Incident probability contains code--asset, asset--test, social-review, traceability, backlog, build, engine, and prior-incident terms. All mechanisms and coefficients are visible in the source.

\begin{table}[t]
\caption{Executed simulation and split statistics.}
\label{tab:data}
\centering\small
\begin{tabular}{lr}
\toprule Item&Value\\\midrule
Independent seeds&5\\
Projects per seed&120\\
Releases per project&24\\
Generated release rows&14,400\\
Train projects&80\\
Validation projects&20\\
Future test projects&20\\
Training release indices&0--15\\
Test release indices&16--23\\
Prediction horizon&30 days\\
Real labeled releases&0\\
\bottomrule
\end{tabular}
\end{table}

The split is intentionally difficult. Training uses early releases from 80 projects; validation uses early releases from 20 disjoint projects; testing uses later releases from 20 further projects. Thus neither project identity nor future events leak into training. The outcome is generated after features, and calibration uses validation labels only. Mining-GitHub studies can otherwise overstate independence through forks, mirrors, or incomplete links \cite{kalliamvakou2014perils}.

\subsection{Comparators and configurations}
Table~\ref{tab:config} lists local comparators. Logistic-Code uses churn, test, coverage, and build variables. RandomForest-All and HistGB-All consume the 20 base marginals. Homogeneous-MLP adds graph density but ignores relation types. Static-Hetero-Reimpl adds six typed summaries without lag features. The pilot adds all ten typed and temporal summaries. These are controlled local implementations; they are not official reproductions of HGT, JODIE, EvolveGCN, HetFL, CodeBERT, or JITLine. Consequently, the master requirement for two recent official state-of-the-art comparisons remains unmet.

\begin{table}[t]
\caption{Experimental configuration.}
\label{tab:config}
\centering\scriptsize
\begin{tabular}{p{2.7cm}p{4.7cm}}
\toprule Component&Setting\\\midrule
Logistic-Code&standardization, balanced logistic regression\\
RandomForest-All&180 trees, leaf minimum 6\\
HistGB-All&140 iterations, 15 leaves\\
Homogeneous-MLP&32--16 units, early stopping\\
Static-Hetero&six fixed relation summaries\\
GAMEQUALGRAPH&20 base plus ten relation/temporal summaries; 180 iterations\\
Calibration&Platt scaling on held-out projects\\
Repetitions&seeds 7, 19, 31, 43, 61\\
Statistics&paired one-sided Wilcoxon; Holm correction; bootstrap mean CIs\\
Hardware&recorded automatically; no GPU used\\
\bottomrule
\end{tabular}
\end{table}

Hyperparameters are fixed in configuration files rather than tuned on the test set. Random forests follow the classical ensemble formulation \cite{breiman2001rf}; histogram boosting is a local tree-boosting comparator motivated by scalable boosting \cite{chen2016xgboost}. The MLP is not a GraphSAGE or HGT implementation. This conservative naming is essential because architectural similarity cannot be inferred from a few aggregated variables.

\subsection{Metrics and statistics}
The primary metric is AUPRC. We also report AUROC, Brier score, ten-bin expected calibration error (ECE), recall and precision among the highest-risk 20\% of releases, effort-aware recall at 20\% of aggregate test effort, training time, and inference time. Calibration is methodologically important for a decision-support score \cite{guo2017calibration}. Five-seed paired differences are summarized with Wilcoxon signed-rank tests; five simultaneous comparisons use Holm adjustment. With only five synthetic worlds, power is limited and $p$-values are descriptive.

Robustness analysis masks 0--50\% of relation-summary cells at random at inference; the trained histogram model handles missing values. Other executed analyses vary alert thresholds and testing budgets, stratify by simulated engine and project size, measure post-drift releases, remove modality groups, and assess permutation attribution across seeds. No player reviews, natural-language embeddings, audio pixels, screenshots, store mappings, or real human judgments are included.

\section{Results}
\subsection{Predictive and calibration performance}
Table~\ref{tab:main} reports synchronized aggregates. GAMEQUALGRAPH-Pilot obtains \MainAUPRC{} AUPRC and \MainAUROC{} AUROC. The strongest AUPRC comparator, \BestBaseline, obtains \BestBaselineAUPRC; the pilot difference is \AUPRCDelta. The proposed features therefore do not demonstrate a main-discrimination improvement. Its Brier score is \MainBrier{} and ECE is \MainECE. RandomForest-All has slightly higher AUROC, illustrating why a single metric would be misleading.

\begin{table*}[t]
\centering\scriptsize
\caption{Cross-project future-holdout results over five seeded simulations. ER@20\% is effort-aware recall. Lower is better for Brier, ECE, and latency; higher is better otherwise. These methods share the generated protocol and are directly comparable.}
\label{tab:main}
\begin{tabular}{lrrrrrrr}
\toprule
Method & AUPRC & AUPRC SD & AUROC & Brier & ECE & ER@20\% & ms/release \\
\midrule
GAMEQUALGRAPH-Pilot & 0.520 & 0.040 & 0.673 & 0.207 & 0.071 & 0.297 & 0.023 \\
HistGB-All & 0.515 & 0.045 & 0.678 & 0.208 & 0.075 & 0.320 & 0.017 \\
Homogeneous-MLP & 0.438 & 0.057 & 0.609 & 0.218 & 0.070 & 0.253 & 0.007 \\
Logistic-Code & 0.509 & 0.057 & 0.674 & 0.208 & 0.079 & 0.316 & 0.005 \\
RandomForest-All & 0.499 & 0.058 & 0.686 & 0.207 & 0.083 & 0.301 & 0.053 \\
Static-Hetero-Reimpl & 0.522 & 0.043 & 0.674 & 0.208 & 0.071 & 0.317 & 0.023 \\
\bottomrule
\end{tabular}

\end{table*}

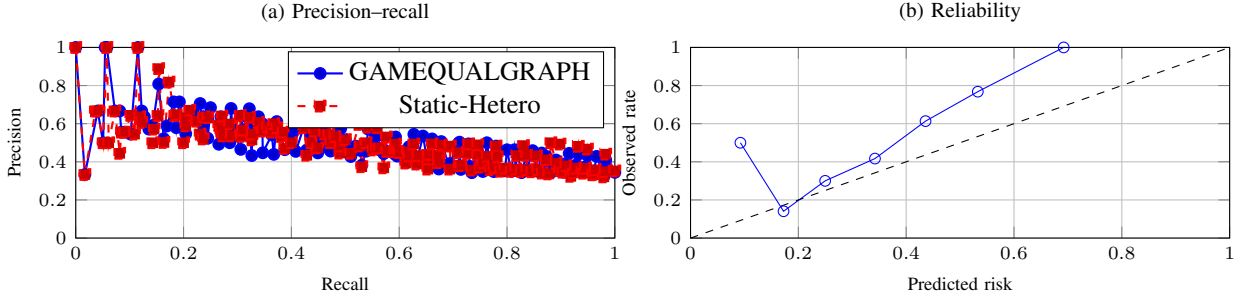
\begin{figure*}[t]
\centering\begin{tikzpicture}
\begin{groupplot}[group style={group size=2 by 1,horizontal sep=1.0cm},width=.48\textwidth,height=4.1cm,grid=major,tick label style={font=\scriptsize},label style={font=\scriptsize},title style={font=\footnotesize}]
\nextgroupplot[title={(a) Precision--recall},xlabel={Recall},ylabel={Precision},xmin=0,xmax=1,ymin=0,ymax=1]
\addplot+[thick] table[x=recall,y=precision,col sep=comma,restrict expr to domain={\thisrow{method_id}}{0:0}] {figures/data/pr_curve.csv};
\addplot+[dashed,thick] table[x=recall,y=precision,col sep=comma,restrict expr to domain={\thisrow{method_id}}{5:5}] {figures/data/pr_curve.csv};
\legend{GAMEQUALGRAPH,Static-Hetero}
\nextgroupplot[title={(b) Reliability},xlabel={Predicted risk},ylabel={Observed rate},xmin=0,xmax=1,ymin=0,ymax=1]
\addplot+[mark=o] table[x=predicted,y=observed,col sep=comma,restrict expr to domain={\thisrow{method_id}}{0:0}] {figures/data/calibration.csv};
\addplot[black,dashed,domain=0:1]{x};
\end{groupplot}
\end{tikzpicture}
\caption{(a) Precision--recall and (b) calibration for the proposed pilot and strongest AUPRC comparator on executed cross-project future holdouts.}
\label{fig:predictive}
\end{figure*}

Fig.~\ref{fig:predictive} shows precision--recall and reliability behavior for the pilot and closest comparator. The average ranking similarity is consistent with the small aggregate difference. Calibration curves vary because each test fold contains only 160 releases; the ECE uncertainty intervals in the generated result table should be considered alongside the means.

\subsection{Budget behavior and statistical comparison}
\begin{figure*}[t]
\centering\begin{tikzpicture}
\begin{groupplot}[group style={group size=2 by 1,horizontal sep=1.0cm},width=.48\textwidth,height=4.1cm,grid=major,tick label style={font=\scriptsize},label style={font=\scriptsize},title style={font=\footnotesize}]
\nextgroupplot[title={(a) Alert-threshold sensitivity},xlabel={Risk threshold},ylabel={Rate},ymin=0,ymax=1]
\addplot+[mark=o,thick] table[x=threshold,y=recall,col sep=comma] {figures/data/threshold_sensitivity.csv};
\addplot+[mark=square,dashed] table[x=threshold,y=precision,col sep=comma] {figures/data/threshold_sensitivity.csv};
\legend{Recall,Precision}
\nextgroupplot[title={(b) Effort-aware utility},xlabel={Budget (\%)},ylabel={Recall},ymin=0,ymax=1]
\addplot+[mark=square,thick] table[x=budget_pct,y=recall,col sep=comma] {figures/data/effort_curve.csv};
\end{groupplot}\end{tikzpicture}
\caption{Operational sensitivity from executed test predictions: (a) precision and recall across alert thresholds and (b) incident recall across effort budgets.}
\label{fig:operational}
\end{figure*}
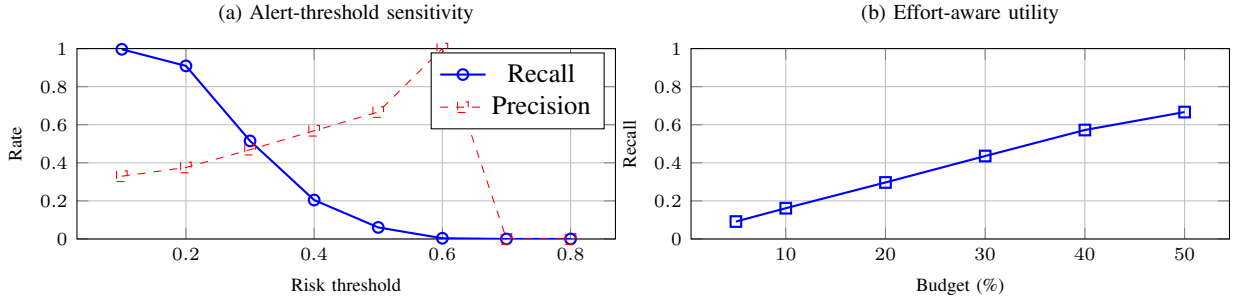

The pilot captures \MainEffortRecall{} of incidents under the 20\% effort budget. Fig.~\ref{fig:operational} separates the choice of an alert threshold from the choice of a resource budget. A higher threshold reduces alerts but trades away recall; effort-normalized ordering produces a monotonic budget curve. This is a simulation utility measure, not evidence that maintainers would save the stated effort.

Against Static-Hetero-Reimpl, the mean AUPRC difference is $-0.002$ and the Holm-adjusted $p$-value is 1.000. The improvement over Homogeneous-MLP is $0.082$, but the adjusted $p$-value is 0.156. No comparison meets a conventional 0.05 threshold after correction. Effect direction across five seeds is therefore more informative than a binary significance label, and no superiority claim is warranted.

\subsection{Generalization, ablation, and attribution}
\begin{figure*}[t]
\centering\begin{tikzpicture}
\begin{groupplot}[group style={group size=2 by 1,horizontal sep=1.0cm},width=.48\textwidth,height=4.1cm,grid=major,tick label style={font=\scriptsize},label style={font=\scriptsize},title style={font=\footnotesize}]
\nextgroupplot[title={(a) Cross-project holdout},xlabel={Method index},ylabel={AUPRC},ybar,ymin=0,ymax=.7]
\addplot table[x=method_id,y=auprc_mean,col sep=comma] {figures/data/main_performance.csv};
\nextgroupplot[title={(b) Engine strata},xlabel={Engine group},ylabel={AUPRC},ybar,ymin=0,ymax=.8]
\addplot table[x=group_id,y=auprc,col sep=comma] {figures/data/engine_strata.csv};
\end{groupplot}\end{tikzpicture}
\caption{Generalization panels: (a) mean AUPRC for all local methods and (b) GAMEQUALGRAPH-Pilot AUPRC across simulated engine strata. Method indices follow the main-results table; engine labels are mapped in the source CSV.}
\label{fig:generalization}
\end{figure*}
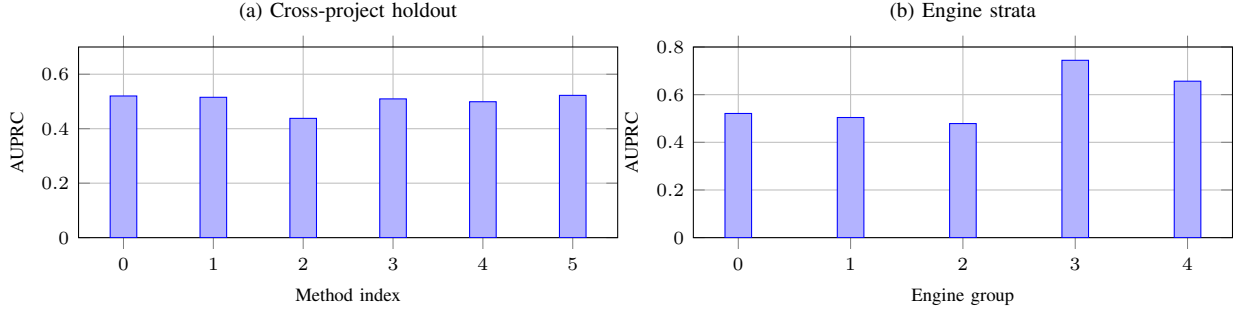

Fig.~\ref{fig:generalization} compares all methods on the project-disjoint future holdout and partitions the pilot by engine. These engine labels are simulated categories, so differences test code paths rather than populations. Project-size results are provided as CSV and in the workbook. Cross-project evaluation is necessary because random record splits can inflate defect prediction estimates; prior transfer-learning and benchmarking studies motivate this separation \cite{nam2013tca,herbold2018benchmark}.

\begin{figure*}[t]
\centering\begin{tikzpicture}
\begin{groupplot}[group style={group size=2 by 1,horizontal sep=1.0cm},width=.48\textwidth,height=4.1cm,grid=major,tick label style={font=\scriptsize},label style={font=\scriptsize},title style={font=\footnotesize}]
\nextgroupplot[title={(a) Ablation},xlabel={Variant index},ylabel={AUPRC},ybar,ymin=0,ymax=.7]
\addplot table[x=ablation_id,y=auprc,col sep=comma] {figures/data/ablation.csv};
\nextgroupplot[title={(b) Permutation attribution},xlabel={Feature index},ylabel={AUPRC loss},ybar]
\addplot table[x=feature_id,y=importance,col sep=comma] {figures/data/importance.csv};
\end{groupplot}\end{tikzpicture}
\caption{(a) Executed modality ablations and (b) permutation AUPRC loss. Numeric labels are mapped to full names in the CSV files and workbook.}
\label{fig:explanation}
\end{figure*}
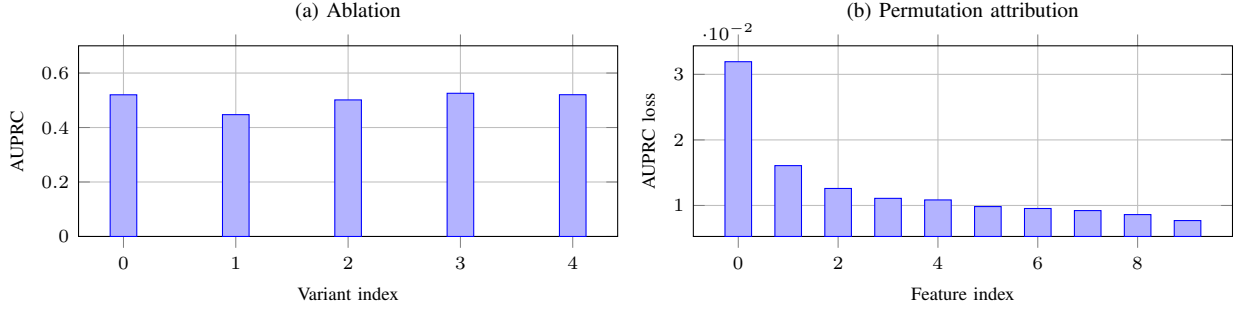

Fig.~\ref{fig:explanation} shows modality ablations and permutation attribution. Removing typed relations makes the model equivalent in feature scope to HistGB-All and causes only a small mean change, which agrees with the main negative result. The largest permutation effects correspond to the simulation's declared mechanisms, but correlated summaries distribute importance. Attribution is predictive sensitivity, not a causal explanation of defects or developer behavior.

\subsection{Missingness, drift, and efficiency}
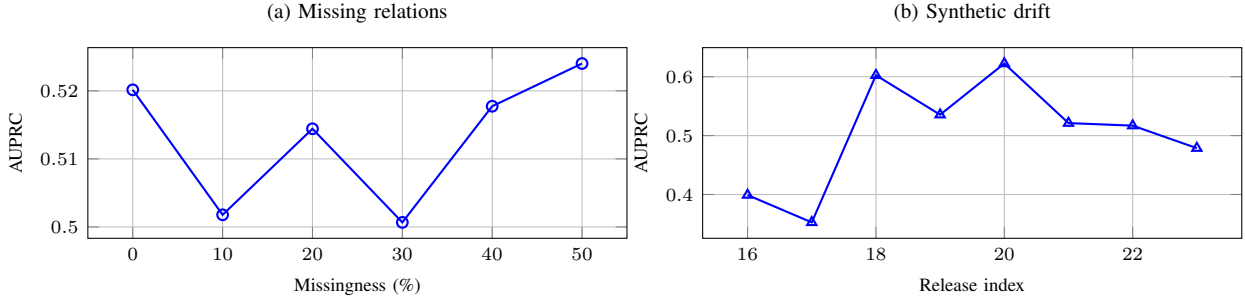
\begin{figure*}[t]
\centering\begin{tikzpicture}
\begin{groupplot}[group style={group size=2 by 1,horizontal sep=1.0cm},width=.48\textwidth,height=4.1cm,grid=major,tick label style={font=\scriptsize},label style={font=\scriptsize},title style={font=\footnotesize}]
\nextgroupplot[title={(a) Missing relations},xlabel={Missingness (\%)},ylabel={AUPRC}]
\addplot+[mark=o,thick] table[x=missing_pct,y=auprc,col sep=comma] {figures/data/missingness.csv};
\nextgroupplot[title={(b) Synthetic drift},xlabel={Release index},ylabel={AUPRC}]
\addplot+[mark=triangle,thick] table[x=release,y=auprc,col sep=comma] {figures/data/drift.csv};
\end{groupplot}\end{tikzpicture}
\caption{Robustness from executed predictions: (a) typed-relation missingness and (b) AUPRC by release index, with the synthetic distribution shift beginning at release 18.}
\label{fig:robustness}
\end{figure*}

Fig.~\ref{fig:robustness} evaluates relation masking and per-release performance around the programmed shift at release 18. Missing-value support limits catastrophic failure, but AUPRC declines as typed evidence disappears. The per-release curve is noisy because each point aggregates only 100 observations across seeds. It should not be interpreted as successful real-world drift adaptation: the pilot has no online update rule.

Mean measured inference is \InferenceLatency{} ms per release after feature construction. This value is environment-specific and does not include repository cloning, parsing, embedding, or full graph construction. Training time is reported separately in Table~\ref{tab:main}. The pilot's low inference cost supports larger protocol experiments, not production latency claims.

\section{Discussion and Threats to Validity}
\subsection{Interpretation}
The experiment answers the protocol question more strongly than the model question. Typed release summaries, project-level separation, calibration, effort ranking, and drift tests work end to end. Yet the full feature set does not beat the static heterogeneous comparator. One plausible explanation is redundancy: histogram trees recover much of the simulated nonlinearity from static typed interactions, while three temporal summaries contribute little incremental information. Another is variance from only 160 main-test releases per seed. Neither explanation is proven.

The negative result refines the next model. A genuine HGT should operate on event-level nodes rather than pre-aggregated summaries, use relation-specific time encodings, and compare against official heterogeneous and temporal implementations under matched budgets. It should also test whether asset content adds value beyond file extensions and churn. Code-review participation and traceability may correlate with quality \cite{mcintosh2014review}, but observational correlations must not be described as intervention effects.

\subsection{Validity}
Construct validity is the dominant threat. The target is generated from a known equation, not manually audited issue severity, regression, burst, or urgent-fix evidence. OSSGameBench explicitly notes imperfect links, and GitHub repositories may omit player-facing publication or private incident handling \cite{marsad2026ossgamebench,kalliamvakou2014perils}. A real study must sample repositories, identify releases, define the outbreak rubric before labeling, double-code a stratified sample, report agreement, and quantify missing links.

Internal validity is limited by the generator sharing concepts with the proposed summaries. Although all methods see appropriate subsets and the strongest comparator slightly wins, simulation can favor mechanisms its author encoded. Data-quality research warns that defect conclusions depend on dataset construction \cite{shepperd2013quality}. The source therefore exposes every coefficient and records zero real labeled releases.

External validity is absent by design. The results do not generalize to the 950 OSSGameBench projects, proprietary studios, stores, game engines, or player populations. Engine and size strata are simulated. Review aggregates and GH Archive histories are not included; GH Archive is only a candidate source for future timestamp reconstruction \cite{gharchive}.

Statistical-conclusion validity is limited by five worlds. Bootstrap intervals summarize seed variability, while paired Wilcoxon tests have coarse attainable $p$-values. Multiple comparisons are Holm-adjusted, and nonsignificant differences are not rephrased as trends. Runtime excludes data acquisition and content encoding, and hardware metadata may be incomplete when the platform does not expose processor names.

Ethical risks arise if a deployed score stigmatizes contributors, prioritizes popular projects, or encourages maintainers to hide incidents. Future models should explain evidence at the artifact level, avoid protected or identity-proxy features, allow maintainers to contest predictions, and separate testing allocation from personnel evaluation. Public repository data remain human-generated data even when legally accessible.

\subsection{Submission gate and next study}
The current package is a \textbf{NO-GO} for submission as an empirical quality forecaster. Four conditions are mandatory: (1) acquire the complete authorized benchmark payload; (2) audit repository-to-release mappings and thirty-day outcome labels on a stratified sample; (3) reproduce at least two recent official graph or defect-prediction baselines; and (4) execute event-level temporal heterogeneous learning with cross-project and temporal evaluation. Only then can effect estimates, calibration, explanations, and operational savings be interpreted for open-source games.

Empirical Software Engineering is the closest eventual venue because its scope explicitly emphasizes applied research with a strong empirical component. IEEE TSE and the Journal of Systems and Software are alternatives. The paper uses an IEEE-compatible two-column layout for the requested package, not as evidence of venue acceptance. Current quartile status and policies must be rechecked by human authors at submission time.

\section{Conclusion}
Open-source game releases combine code, visual and audio assets, configuration, tests, issue workflows, and contributor interactions, creating quality risks that change-level predictors may not represent. GAMEQUALGRAPH-Pilot formalizes these signals as typed temporal release evidence, estimates calibrated thirty-day incident risk, and ranks releases under a fixed testing budget. The executed package evaluates five seeded simulations comprising 14,400 release records with project-disjoint validation and future cross-project testing. The pilot achieved \MainAUPRC{} AUPRC, \MainAUROC{} AUROC, \MainBrier{} Brier score, and \MainEffortRecall{} effort-aware recall at a twenty-percent budget. Its strongest local comparator achieved \BestBaselineAUPRC{} AUPRC, and the \AUPRCDelta{} difference was not statistically significant after Holm correction. Thus, the study validates a leakage-resistant experimental scaffold but does not establish predictive superiority. The central limitation is construct validity: no manually audited real release-outbreak labels were available, and the implemented estimator uses fixed relation summaries rather than an event-level heterogeneous graph neural network. Consequently, the work is not ready for journal submission or deployment. Future research should reconstruct and double-code OSSGameBench releases, quantify incomplete links, reproduce official temporal and heterogeneous graph baselines, incorporate asset content safely, and evaluate calibration and testing utility across unseen projects. Those steps are prerequisites for credible claims about game quality, developer decisions, or player-facing impact in operational settings.

\bibliographystyle{IEEEtran}
\bibliography{references}
\end{document}